# Tragic and Comical Networks
# Clustering Dramatic Genres According to Structural Properties

| Botond Szemes | Bence Vida |
|---|---|
| Research Centre for the Humanities<br>Institute for Literary Studies Budapest<br>szemes.botond@abtk.hu | Eötvös Loránd University<br>vida.bence@btk.elte.hu |

## Abstract

There is a growing tradition in the joint field of network studies and drama history that produces interpretations from the character networks of the plays. The potential of such an interpretation is that the diagrams provide a different representation of the relationships between characters as compared to reading the text or watching the performance. Our aim is to create a method that is able to cluster texts with similar structures on the basis of the play's well-interpretable and simple properties, independent from the number of characters in the drama, or in other words, the size of the network. Finding these features is the most important part of our research, as well as establishing the appropriate statistical procedure to calculate the similarities between the texts. Our data was downloaded from the DraCor database and analyzed in R (we use the GerDracor and the ShakeDraCor sub-collection). We want to propose a robust method based on the distribution of words among characters; distribution of characters in scenes; average length of speech acts; or character-specific and macro-level network properties such as clusterization coefficient and network density. Based on these metrics a supervised classification procedure is applied to the sub-collections to classify comedies and tragedies using the Support Vector Machine (SVM) method. Our research shows that this approach can also produce reliable results on a small sample size.

## 1. Introduction

Do the various dramatic traditions and genres have any statistically verifiable structural features in common; i.e., are there any features of dramatic structure that guide or even presuppose the process of creating plays irrespective of the authors' individual style? Considering how interesting these questions are, the amount of debate it generates in the field of quantitative drama analysis should come as no surprise (the earliest attempt is by Boris Yarkho, see: Gasparov 2016). Indeed, an exploration of this problem has the potential to offer not only a novel approach to drama and theater studies but also access to the broader mechanisms of culture. To that end, the question might be reformulated in the following way: how are the relationships between people or other actors arranged according to the type of story; or, vice versa, what types of stories does each arrangement make possible? An adequate methodology needs to be designed to answer these questions and identify the characteristics that make a comedy a comedy or a tragedy a tragedy. This should provide a new perspective to this comparison, especially as earlier research distinguished comedies from tragedies mainly based on their thematic differences, such as the extent to which historicity determines the setting and the period the events take place in, the emphasis on moral issues, the social status and speech patterns of the characters, or the conclusion of the plot. (Asmuth 2016; for a theme-based classification of dramatic genres with topic modelling, see: Schöch 2017).

The present paper suggests that the primary key to answering these questions is the character networks of plays, which allows the plot (Moretti 2011), interpersonal relations and societal world (Algee-Hewitt 2017), and/or dramatic form (Trilcke and Fischer 2018, and also see their study in the present volume) of a given play to be viewed simultaneously on the plane of a surface instead of in the process of reading a play or watching a performance, which inevitably unfold over time. In our study we use the concept of *structure*, by which we mean the system of relations of characters in the broadest sense (and not only network relations, as will be explained later). This concept is closest to *form*, but it is also related to (and thus brings together) the other approaches mentioned – e.g. to plot: the analysis of structure also deals with the relationships between characters as a result of the progression of the plot (for an analysis of one aspect of this, see section 4.1.) Moreover, following the considerations and methods of SNA (Social Network Analysis), character networks can also be seen as structural representations of the contemporary cultural-historical context of a given drama; and thus plays can be understood as the author's – not necessarily intentional – attempt to model a society. (Stiller, J., D. Nettle, and R. I. M. Dunbar. 2003; Stiller, J. and M. Hudson 2005).

Furthermore, these networks, beyond being a cognitive tool to get a comprehensive overview of a text, can be also compared and classified statistically. The main implication of this is that the approach and terminology of graph theory can be applied to character networks drawn from the text; in our case, the characters represent the nodes and the edges are created by characters appearing in the same scene (in the case of weighted graphs, the more often the characters appear together, the thicker the edge between them). This allows the use of several numerical measures to specify the properties of the network. While specific analyses conducted on a small population sample and at the level of individuals are better suited to the SNA approach (Freeman 2004; Borgatti et al. 2009), we endeavored to use or create numerical measures that describe the plays in their entirety, so that they can be compared to each other. Our hypothesis is that the plays' plot establishes a network based on interactions (dialogue and action), which the author subjects to the criteria of the genre he or she chooses.

Earlier similar research also drew mainly on character networks when trying to explore the differences and similarities between dramatic genres (comedies and tragedies). These projects however suffer from two crucial problems that call into question the usefulness of their results. First, most papers rely heavily on the size of the networks to classify genres, which instead of capturing the structural design of networks, simply compares the number of characters in the plays.[1] In what follows, an attempt will be made to present a method that does not take the size of the networks into account, i.e., it will be based exclusively on the system of relations between characters, and through this on the plays' dramatic structure. The other pitfall to be avoided is the application of overly complicated mathematical procedures to investigate similarities between networks (e.g. Cameron 2020, Tsitsulin et al). The problem with this approach is that the results obtained this way cannot be attributed to real properties of the plays; such results would only show whether two works have a similar structure. By contrast, our goal is not simply to confirm that there is such a similarity but also to explain what this similarity consists of. Accordingly, the analyses below will be founded on data that truly describes how these works are structured (i.e. how they elaborate relationships and what kind of hierarchy they create between the elements): first, size-independent numerical measures of character networks; and second, values that we generated pertaining mainly to the distribution of speech and stage time among characters, which thus can describe relations in a different way, in terms of *ratios* (the most prominent and earliest representative of this approach is, again, Boris Yarkho, cf. Yarkho 2019).

## 2. Corpora

To explore the issue, we needed a large number of plays available in a structured form, which was provided by the database of the DraCor Project (Fischer et al. 2019). This project, a product of an international collaboration, acts as the corpora of the dramatic literature of various nations. More than a mere collection of texts, it also supplies important tools for analyzing these texts, most notably the automatic mapping of character networks and access to the data behind them. This makes it possible to use network theory measures for describing the relationships in the works, which in turn lays the foundation for a quantitative comparison of the structure of plays.[2] For the present research the 37 plays of the

1 See Trilcke (2015), Grandjean (2015), and Shukla et. al. (2018). In Mark Algee-Hewitt's paper (2017), which is a milestone in quantitative drama analysis, it is unclear how independent the numerical measures are from the number of characters.

2 DraCor provides a standardized framework that can be used across all languages, with its core being the dramatic texts encoded in TEI XML format. In the process of coding, the annotators identify the plays' structural elements, the acts and scenes; label the characters separately; and also indicate which characters appear in a given act or scene. This helped create the networks that were extracted for the present study (using the simple network visualization tool of the interface) to complement the lists of characters available in .csv format and downloadable one by one for each play. Therefore, the raw data contains the structure of each play,

Shakespeare collection and the "GerDracor" database of German plays were used. Beyond the fact that the Shakespeare collection attracts special attention in quantitative drama analysis, like in so many other fields, it also provides a well-defined sub-corpus on which to carry out tests, and its small size allows for the interpretation of the results on specific texts. GerDracor, by contrast, supplies a large number of texts, which is ideal for statistical analysis. At the moment, this database stores 591 plays, but only those with a clear genre designation (comedy or tragedy)[3] and more than five characters and more than two scenes will be analyzed below; in all other cases, character networks cannot be said to be indicative of dramatic structure. Applying these reductions, the collection finally contains 253 dramas. History plays from the Shakespeare corpus were included in the tragedy genre, since Shakespeare criticism suggests that a strong similarity can be detected between these two genres (Kott 1974; Bate 1997). Therefore, we used a binary classification (comedy vs non-comedy) in this case as well. The two corpora have different characteristics because of the divergences in their size and diversity. For example, as we will see, the number of characters in Shakespeare's plays correlates more strongly with other structural features of the genres, whereas in GerDarcor the metrics are more size-independent – although it is important to bear in mind that GerDracor is much more spread out in terms of authors and time, and thus represents a less coherent unity. In what follows, we take the results for the GerDracor as a baseline, as it shows statistically more reliable results in a larger scale, but the variation in the corpora also indicates that it is always worthwhile to design the method according to the corpora under study.

### 3. Numerical Measures

To determine the size-independent but distinctive properties of plays, the following numerical measures were considered, and in some cases developed for this study. The first four measures are considered standard in SNA studies, and they are featured in most papers dealing with quantitative drama analysis.

*Average Clustering Coefficient*: This measure indicates the proportion of nodes connecting to distinct sub-units that have a high number of edges in a given network. The small-world principle states that most social networks are highly clustered (Watts and Strogatz 1998).[4]

*Density*: A ratio of the number of actual edges in a network to the number of possible edges. The denser the network, the more connections can be observed in it, which provides

---

along with unique identifiers of characters appearing in the same scene. In the present study, connections were established between those characters that appear at the same time in the same scene according to the annotated list of characters.

3 The metadata in GerDracor's corpus also specify the so-called "normalized genre tags," which standardize the different designations that refer to the same genre tradition (e.g., *Lustspiel* and *Komödie*).

4 Stiller et al. (2003) demonstrate that a large cast of characters increases the clustering of the play's character network, which can thus be described as a small world. Clustering can also be used to identify weak ties, an important subject in SNA that measures the resilience of social networks and the effectiveness of information diffusion. (see Granovetter 1983; Albert et al. 2000) information on the network's complexity and type. For example, a democratic network (informal group of friends) has a higher density than a hierarchical one (chain of command).

*Average Path Length*: In the network a path can be drawn between any two nodes through the edges. This measure shows the average length of the shortest path that can be drawn between any two nodes (the path that touches the least number of nodes).

*Diameter*: Finding the shortest path between nodes in a network is crucial to defining its basic properties. The diameter represents the longest of these paths, which is also the distance between the furthest nodes.

*Maximum Betweenness*: The betweenness centrality for node A indicates the number of possible shortest paths that connect any two nodes of the network and touch A, relative to all possible shortest paths. In other words, betweenness is the centrality measure that assesses the mediating role of a given node, and thus answers the question: how many other nodes are involved in information transfer or path formation between nodes. The Maximum Betweenness indicates the maximum value of these centralities. The measure is particularly useful for identifying nodes that may act as bridges between clusters in a network.[5]

*Average Degree – Max Degree Ratio:* Degree connotes the number of nodes that any given node is connected to via edges. From this number, the average and maximum degree for the network as a whole can be calculated. This is done so that statements could be made that pertain to the network itself. This measure therefore gives the proportion of the average degree number relative to the maximum degree number, and in a sense resembles density. The extent to which this measure can be considered an independent characteristic that describes networks from a unique point of view will be discussed later.

*Maximum Degree – Number of Characters Ratio:* In this case to get the total number of characters correct, a value of one has to be subtracted from the actual number of characters, since a node does not interact with itself. The ratio shows what percentage of the entire cast the node with the highest degree interacts with.

*High Speech, Medium Speech, Low Speech:* These three measures refer to the proportion of characters who speak "much", "little", or "a moderate amount" during the play. This was calculated according to the procedure outlined in the Stanford Literary Lab's *Pamphlet 14* (Kanatova et. al 2017), which used the k-means clustering procedure to classify movies into 3 groups (k=3) based on the amount of analepses and prolepses they contained, thus distinguishing works that scored "high", "medium" and "low" from each other (i.e., the group with the highest mean value constituted the "high" category and so on.) In the present paper, k-means clustering was employed (k=3, likewise) to distinguish between characters in plays who are not just members of a group in the stage (for which DraCor uses the annotation "isGroup = True", e.g. "street vendors"). The three groups were then ranked according to their mean value, and the number of characters in each category was then adjusted to the total number of non-group characters in the play. This ratio makes it possible

5 This measure was also used by Mark Algee-Hewitt (2017) in his research.

to compare plays with different numbers of characters, since it shows not only the number of characters who speak a lot/medium/less often in a text, but also the proportion of characters who do so, which essentially gives us an idea of the distribution of linguistic complexity among characters.

*High Weighted Degree, Medium Weighted Degree, Low Weighted Degree:* The same method was employed to calculate the weighted degree of characters as nodes. Here, the proportion of characters with many, some or few connections in the network was examined. Since the result is always adjusted to the total number of characters, this measure is also size-independent, which allows us to compare works with different numbers of characters. Our previous measures showed that the metrics of weighted degree (which takes into account not only the number of edges leading into a node but also their "weight", in this case the frequency of appearing together in the same scene) better captures the differences between genres and the specific characteristics of plays than using only the degree for the calculations.

*Average Character Speech:* This shows the average number of words spoken by a character in a play. Only characters who speak more than 10 times were considered.

*Average Character Per Scene:* First, the average number of characters in one scene is calculated. To make texts of different lengths comparable, this number is divided by the total number of scenes.

*Number of Connected Components:* The number of groups in the network within which any node is connected to any other node by edges (Minimum = 1).

Before presenting the results, a discussion on how these measures relate to each other - whether a correlation can be detected between them - is in order. Are these measures independent of each other and of the size of the networks? The upper part in the correlation matrix of Fig1 created from the values of 253 plays of GerDraCor shows these connections, while the lower part shows the results based on the Shakespeare collection. In the upper triangle (which serves as baseline to our study) the "number of connected components" is strongly correlated with the number of speakers (i.e., the size of the network), which is hardly surprising, since it is almost exclusively found in the case of plays with a high number of characters, which means the network can separate into multiple unconnected units. Thus, this measure was not used in subsequent analyses. In addition the correlation between density, average path length and average degree – maximum degree ratio is very strong (>0.9); therefore the latter two measures were also excluded from the study, since density is one of the most important and informative properties of character networks (so it was deemed a more characteristic, more descriptive measure than the other two; and furthermore the exclusions make sure that the strong correlation between diameter and average path length [0.88] do not influence the subsequent results.) Although there are still some pairs of values that are correlated (e.g., density – average character in one scene), the relationship between them is not exclusive, i.e., they cannot be said to describe the same phenomenon or phenomena that are in a deterministic relationship (see Fig.2). Regardless, it should be noted that they reinforce each other and have a joint effect in the clustering process (see Shukla et al. 2018).

The lower triangle of Fig2. shows that a smaller corpus of one author is organized in a different way. Here we can observe a stronger correlation between density and the size of the networks.Perhaps this together with the fact that almost exclusively Shakespeare's texts were studied previously in similar research, made it important to also involve the number of characters into the classification process. However, it seems that for measurements on a larger scale, the features are less dependent on the number of characters, which would also play a less important role in the classification (see later).

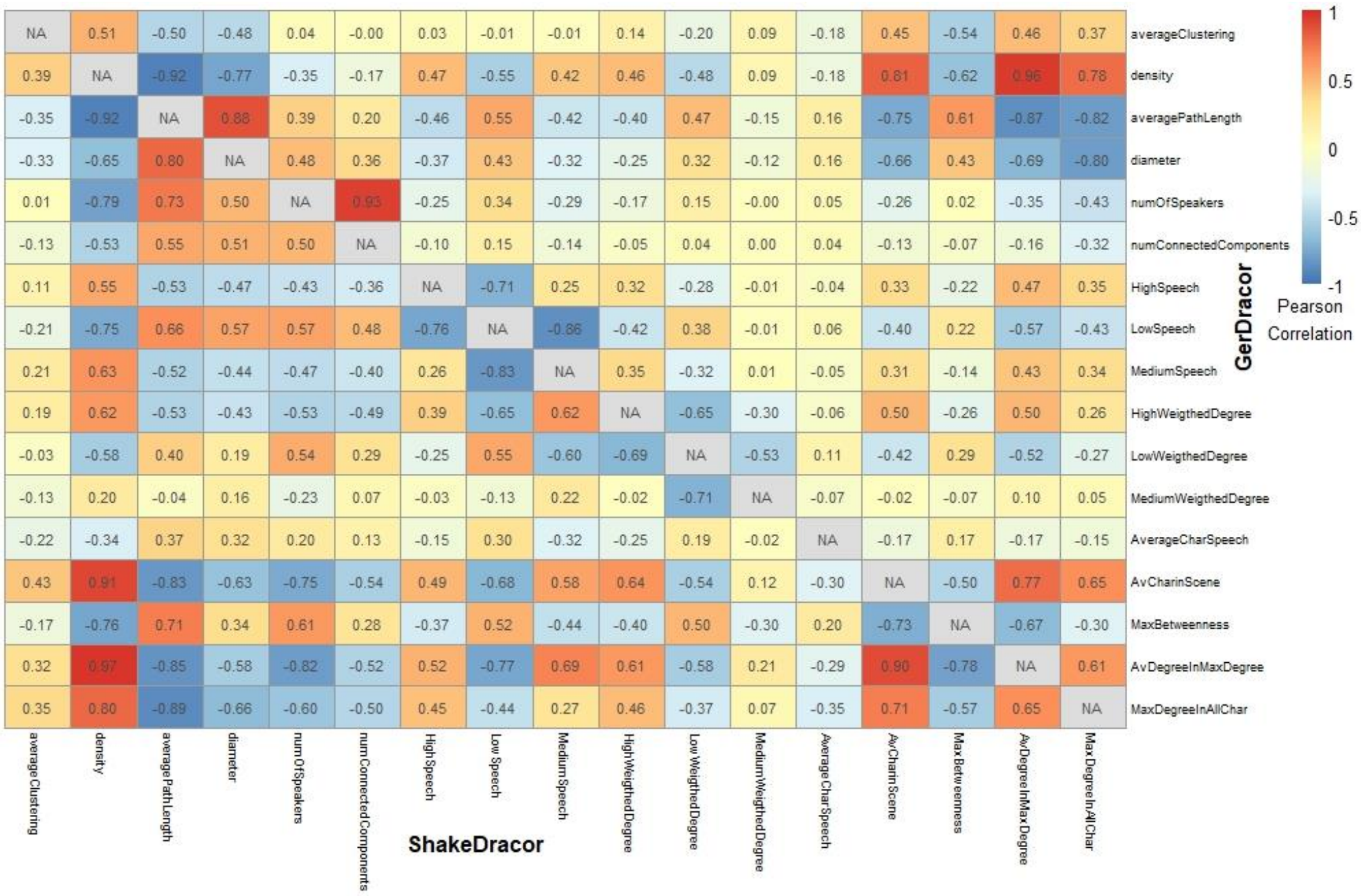

Fig 1. Correlation matrix of the values used in the research. Upper triangle based on the GerDracor, lower triangle based on the ShakeDarcor database

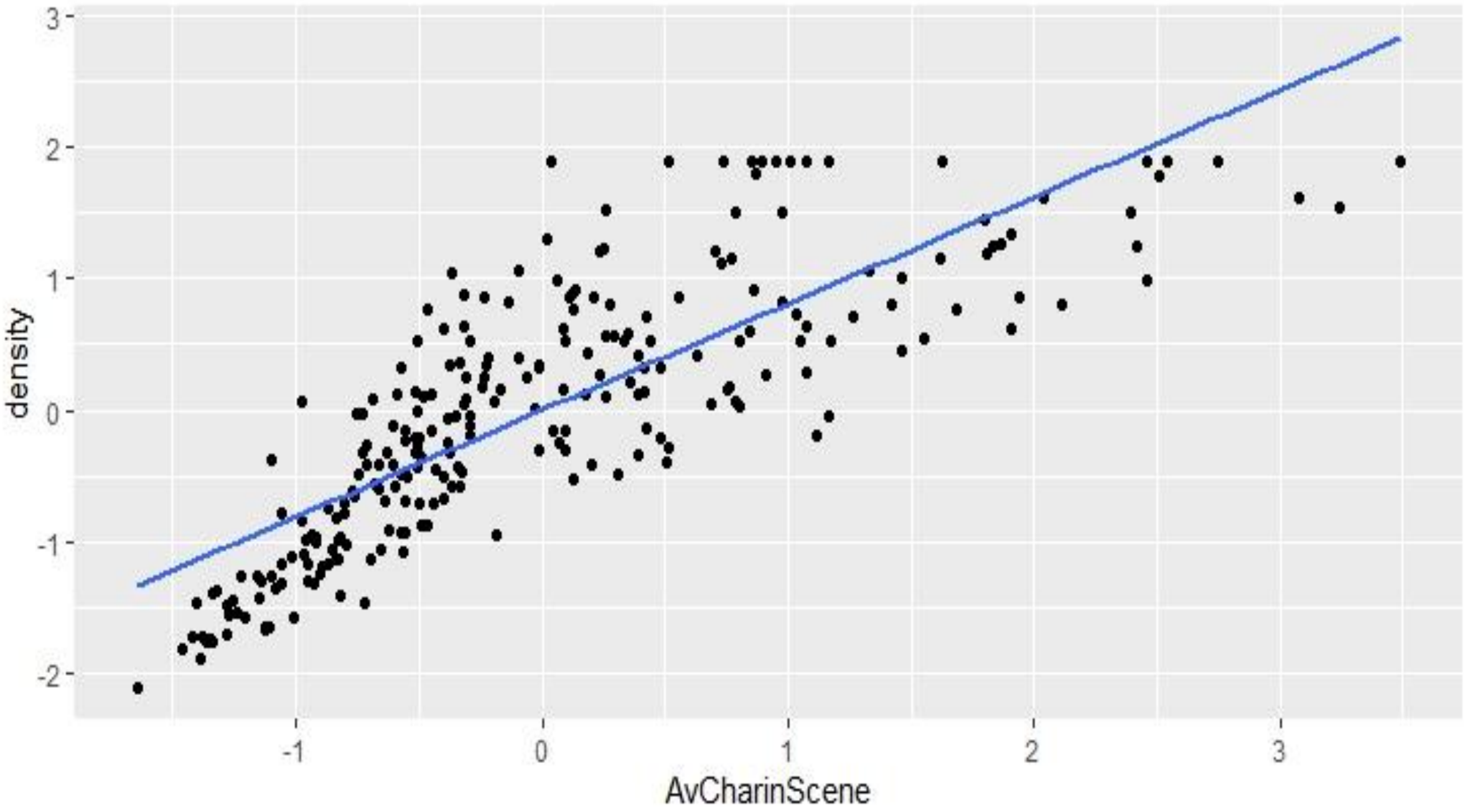

Fig 2. Correlation between density and average character per scene based on the GerDracor database. Normalized values presented (z-scores).

The differences between the 13 variables (which remained after taking into account the correlations based on GerDracor) across genres are shown in Fig4 and can be considered significant based on the Wilcoxon Rank Sum Test, except for the category of the proportion of the medium weighted degree (this test was chosen instead of the t-test because the variables do not follow a normal distribution). The calculations were then performed using the normalized values (z-score) of the 13 variables for the corpus in question so that differences of different scales would be given equal weight when determining genre groups.

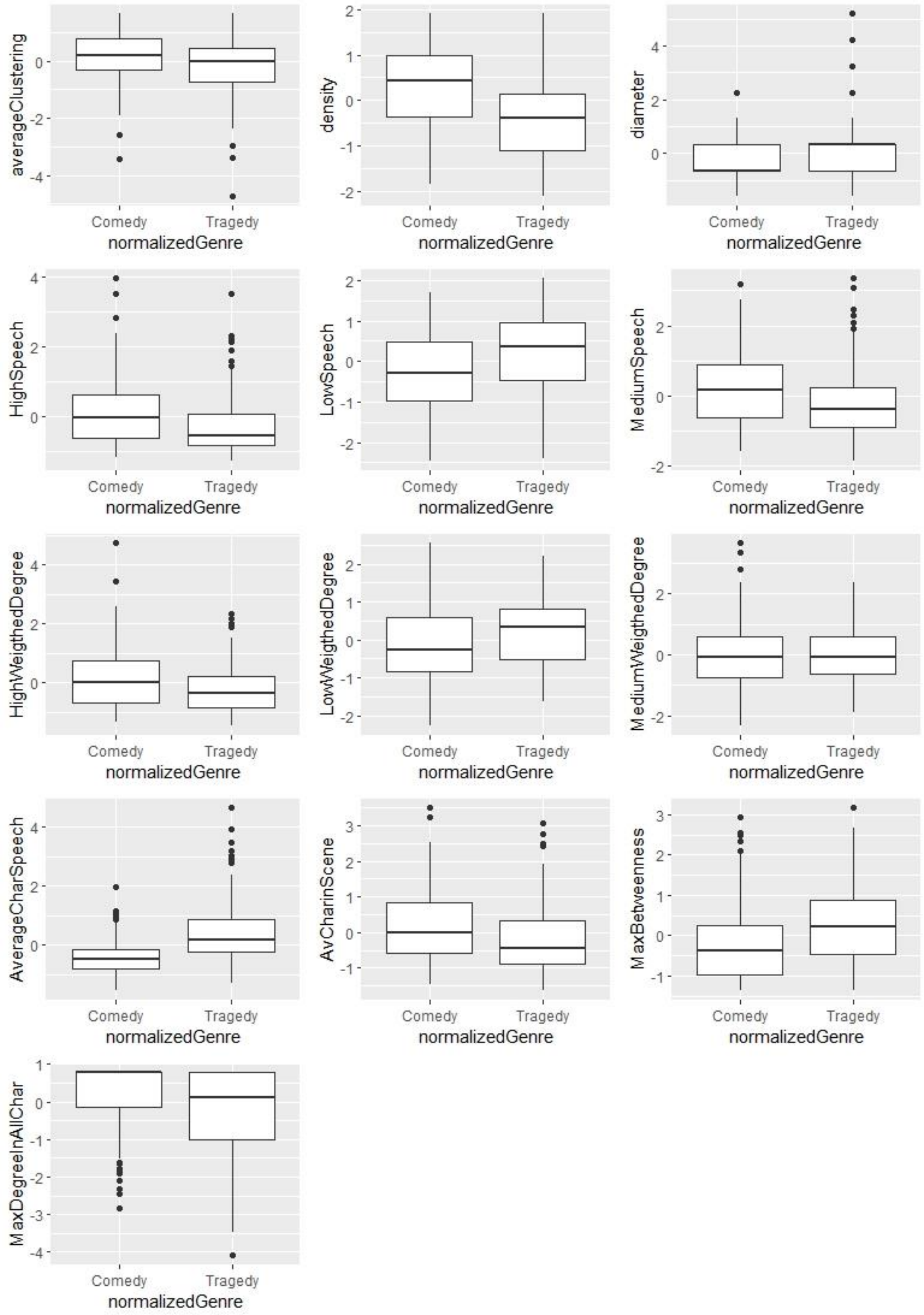

Fig3. Differences in the features by genre based on the GerDracor database. Normalized values presented (z-scores).

## 4. Results

As a first experiment a supervised classification procedure was applied to the GerDracor sub-collection using the leave-one-out (LOO) method. In other words, one of the 253 dramas was always withheld at the beginning of the experiment, and then the remaining 252 works were divided into comedies and tragedies using the Support Vector Machine (SVM) classification procedure. This procedure can be viewed as if each play were a data point in a 13-dimensional space, where the SVM “drew” the hyperplane that best separates the categories (since we used a linear kernel, this plane is created without curvature). We then examined which side of the space (comedy/tragedy) the withheld work would lie on based on the values assigned to the variables; and assessed the correctness of the binary classification on this basis. This withholding was repeated with every single play (separately for tragedies and comedies). The accuracy of the classification was finally gauged by the average of these numbers, which is represented by the mean accuracy score – which in this binary case is the same as recall.

Since the corpus contained an uneven number of tragedies and comedies, the probability for random classification was not 0.5 but 0.54 (136/253) for comedies and 0.46 (117/253) for tragedies. The experiment produced an accuracy rate of 0.82 (F1 score = 0.77, precision = 0.73, recall = 0.82) for comedies and 0.65 (F1 score = 0.7, precision = 0.76, recall = 0.65) for tragedies, which suggests that the model is more accurate than random classification and its performance is quite good, but far from flawless. If we include the number of characters as a feature to the model, the performance is roughly the same: 0,83 mean accuracy rate for comedies and 0,64 for tragedies; and furthermore this feature proved to be the second least important in the classification after testing the role the number of characters plays in the model (methods described later). Which means that in this case the size of the network is far less relevant in determining the genre of a play than previous studies suggested. Moreover, the promising results of the original model imply that the properties examined here can indeed confirm the existence of a “genre fingerprint” that shapes the dramatic structure of tragedies and comedies. However, it is worth pointing out this fingerprint does not determine structure categorically but rather provides a scheme from which individual authors often deviate.

Fig4, generated by principal component analysis (PCA), attests to the same phenomenon: there is no striking difference between comedies and tragedies in their structural properties, but one can detect a tendency of the comedies to gravitate towards the left side of the figure and the tragedies to the right side. The outliers are also positioned along the same lines. Furthermore, the figure shows which side each variable can influence on the plane generated by the PCA’s reduction of dimensions, i.e., how the variables affect the location of data points. The possibility of such a visualization and the satisfactory results of the SVM require us to keep all the variables in the model – although the RFE (Recursive Feature Elimination) procedure for feature selection[6] suggests leaving the features of High,

6 The RFE works in two steps: first it calculates the importance of the features (based on the role they play in Medium and Low Medium Weighted Degree out of the research,as they play a less prominent role in distinguishing genres. But the difference between the accuracy scores of the remaining 10 features (0,72) and of the whole set (0,7) is not as relevant as the importance of taking into account the most probable aspects in defining the specificities of the genres – and thus obtaining a ground for more complex interpretations. According to the PCA graph, for example, most of the comedies are marked by dense character networks and scenes with multiple characters; they have more characters who appear with many characters in the same scene; more characters who talk a lot throughout the play; and finally the character connected with the highest number of characters can be linked with a high percentage of the entire cast. By contrast, in prototypical tragedies there are more characters who speak very little and appear with few others in the same scene, but the length of utterances per person is longer (i.e., there are more monologues), the diameter of the network is larger (i.e., a character might mediate between two other characters who do not interact with each other directly), and one or two characters with a special ‘connecting’ function are more often responsible for the stability of the network, i.e. they are typically the ones through whom information is passed between the different subgroups (maximum betweenness).

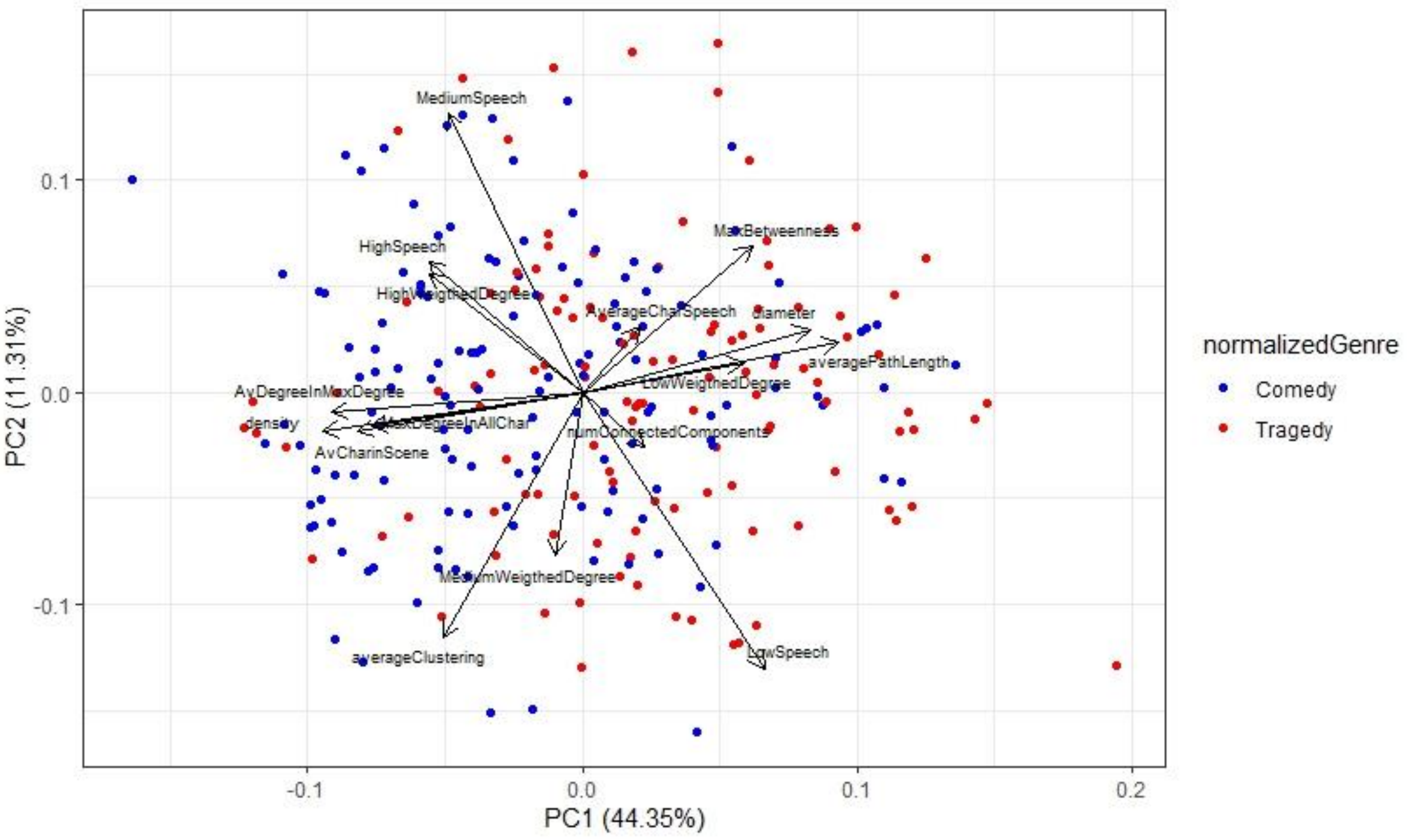

Fig4. Principal Component Analysis based on the measured features - 253 dramas from GerDraCor.

the classification), then it removes the least important ones iteratively and calculates the accuracy results for each scenario.

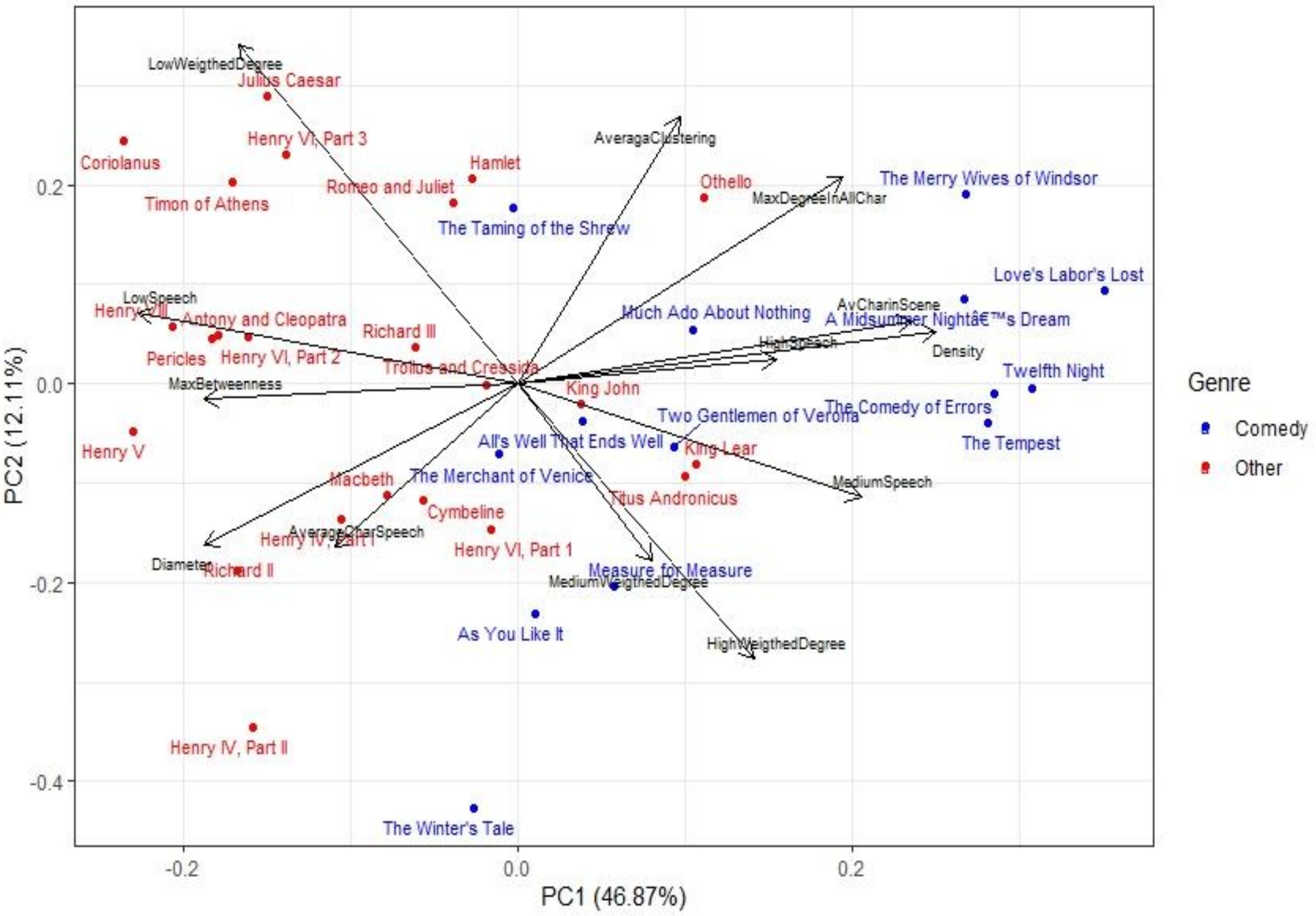

Fig5. Principal Component Analysis based on the measured features - 37 dramas of the Shakespeare corpus.

An examination of the Shakespeare corpus with the same setup yields similar results, which can more easily be put into practice at the level of individual dramas. Here, when included, the number of characters becomes the most important feature of the classification, which has an accuracy of 0,71 for comedies (random at 0,37) and 0,83 for tragedies (random at 0,62).Without the number of characters, tragedies stay at 0,83, however comedies decline to 0,6; which also shows the importance of the network size in this case. But we ought to also bear in mind that the small size of the Shakespeare corpus provides less reliable results than the GerDracor corpora. More interesting is the PCA diagram (Fig5), that shows a clearer distinction between genres than the German sub-corpus, even though the variables exert their influence on very similar areas. In the figure, the prototypical comedies can be easily identified (*The Merry Wives of Windsor, Love's Labour's Lost, A Midsummer Night's Dream*, *Twelfth Night, or What You Will, The Comedy of Errors* and, somewhat surprisingly, *The Tempest*, which most scholars would argue is a blend of genres) as well as the non-comedies (incidentally, their location on the figure confirms our hypothesis that history plays and tragedies do not differ, at least not in terms of their structure). Works that depart from the schema are also conspicuous in the figure. The reason *The Taming of the Shrew* falls close to tragedies most probably lies in its use of frame scenes, since that increases the size of networks, decreases its density, and reduces the number of characters who are connected to each other and speak frequently or moderately often – all of which evoke the structure of tragedies.

*The Winter's Tale* also unmistakably stands apart from the rest of the plays, which can be related to its elusion of distinct genres; accordingly, even its classification as a comedy is debatable. More interesting are the tragedies that appear among the comedies, such as *Titus Andronicus*, *King Lear*, and *Othello*. *Titus Andronicus*, one of Shakespeare's early plays, is a long revenge tragedy in barely structured acts and scenes featuring a high number of minor characters.[7] On several occasions, the ill-structured scenes include situations in which sub-groups of characters exit or enter the stage without interacting with each other. Unfortunately, since our model relies on the annotations from DraCor, it does not detect this movement, as it assumes a connection between all the characters in a scene. In the future, a solution to deal with such situations could be to introduce a method that focuses on such subgroups and the two-way nature of interpersonal dialogues; such a method would indicate a transition even within a scene if the set of speakers changes (see Labatut et al. 2019; Park et al. 2012). However, the case is different with the other two tragedies. There the tragic plot unfolds in a framework that would also be suitable for comedies. Ultimately, the tragic nature of these plays stems precisely from the fact that the not particularly tragic arrangement of social relations (i.e., the exchange of information between characters appears not to be overseen by one or two characters) still leads to fatal mistakes and punishment. The plot of *Othello* is set almost entirely within one subgroup, the household of the main characters is shattered only as a culmination of the dramatic twists in the last scenes. Shakespeare's trick is to pick a domestic environment as a setting for what is essentially a tragedy played out in the public sphere and motivated by a lust for political power. The conflict upsetting the social order thus remains hidden throughout, lurking in the depths of long dialogues, only to suddenly tear apart the tense structure of the surrounding network. Part of the reason Iago can carry out his plan without a problem is that he divides the characters' trust-based, prototypically non-tragic system of relations into subgroups that can be played against each other.[8] A similar dramatic device is at work in *King Lear*. A number of elements evoke the world of comedies on a thematic level: the populous family where the king is not at the center (at least according to the centrality measure, which is also low for the other characters); the figure of the Fool who probes issues associated with language philosophy and ontology in playful prattles; and the other characters, who come to resemble him more and more (especially Kent, who pretends to be a fool); the misunderstandings caused by the disguises; and the forest as a setting, which in comedies is the space where one can indulge in desires repressed in the social sphere. The measuring we conducted confirm the affinity with comedies. The structure of King Lear promises a comedy, albeit not a prototypical one: the fact that it does not follow that path is the source of its tragedy. Lear is

7 Shukla et. al. (2018) also struggles with classifying this work; they treat it as a comedy.
8 Stiller, J. and M. Hudson (2005) describe this phenomenon from a different perspective by focusing on edges in their analysis of *Othello*: Iago consistently tries to manipulate those partners of his who, like himself, are in a 'connectional' role (Cassio, Othello), in such a way as to slowly take control of the flow of information, thus eroding the trust of the others and the stability of the network. His work succeeds at the climax of the play, as the network falls apart from within in the escalating conflict.

unable to "live happily ever after" even though his prospects seem good, as the play opens with what comedies usually end with: a marriage. It ends, however, in the traditional setting of tragedies – among graves (Géher 1999). Thus, the play ambitiously traces an arc from comedy to tragedy, raising questions about the conventions of both.

This leads us to a question of the connection between plot development and the metrics used in this study. This is the question captured by an often-cited observation ascribed to Bernard Shaw, according to whom the two genres are distinguished in Shakespeare according to whether they end with a wedding (where all the characters are present, so the network of characters becomes denser) or the death of the characters (causing a less dense network). (Trilcke et al. 2015) From this angle, *King Lear* could be understood as transitioning from one genre to the other even on a structural level, because both plot points are emphatically present in it. But how can we model the connection computationally?

### 4.1. Results without the last act

To put this question to the test, we repeated our earlier measurements first for the Shakespeare corpus (because this is the domain where the question most often arises), but this time the character network of the plays was based only on the relational structure of four acts – i.e., one act was always disregarded. We examined how this would change the measures describing the network as a whole (in this case density) for tragedies and comedies, and whether the difference between the two remains significant. First and foremost, we were interested in the role of the last act, which is of particular importance to the plot (data for the other acts available in the project's GitHub repository).

The importance of the last act lies in the fact that, according to some researchers, tragedies typically end with the isolation of the protagonist, who gradually becomes a lonely figure and suffers losses, whereas the conclusion of comedies is a large crowd scene on stage (Smith 2009). James J. Lee and Jason Lee, by contrast, seek to demonstrate that the final scene is always a crowd scene, since the conclusion of tragedies also requires multiple characters, and the real difference hinges on the type of the relationship between them (Lee and Lee 2017). In their view, the tragic conclusion is caused precisely by the close and dense arrangement of the characters, which allows the conflicts to unfold. If true, this stresses the importance of the nodes that create weak links, since it is their role to disrupt the stability of the network: it is because of them that the initial network of the play turns into that of the denouement (for a similar interpretation about the *Hamlet* and Horatio's role see: Moretti, 2011).[9] This theory is not confirmed by the results of the measurements, which rather support the distinction made in Shakespeare criticism. Indeed, removing the last act on average decreased the density of the networks in comedies and increased it in tragedies (see Fig6, which shows the effect of the last act; by simply subtracting the score without the final act from the total density, a positive number indicates that the measure is higher when the

9 This argument can be linked to an idea that has been put forward from time to time in Shakespeare scholarship, which argues that the overarching motif of the Shakespeare corpus is the destruction of the established social order. Network analysis shows that this destruction is not self-serving or anarchic but transformative, with the key agents of the new structure always doing their work before the audience, bringing the outcome closer, gossip by gossip, intrigue by intrigue.

conclusion of the plot is included). This means that in the Shakespeare corpus the structure of the networks is indeed influenced by the genre-specific closure of the plot. This is not apparent in the case of GerDracor, but for both corpora, on average only the last act has a positive effect on density in comedies (Fig7) – which can be explained also by the fact that new characters are rarely introduced in the last acts of comedies. But at the same time, a nuance in the picture is that the difference in density also remains significant without the last act based on the Wilcoxon-test (p-value for the whole drama: 0.0003702; p-value without the last acts: 0.00002745). This supports the idea that the structural properties of the plays (at least in the Shakespeare corpus) are influenced by the development of the plot, although not exclusively. The plot is only one component, though an essential one, in the formation of the relationships between the characters.

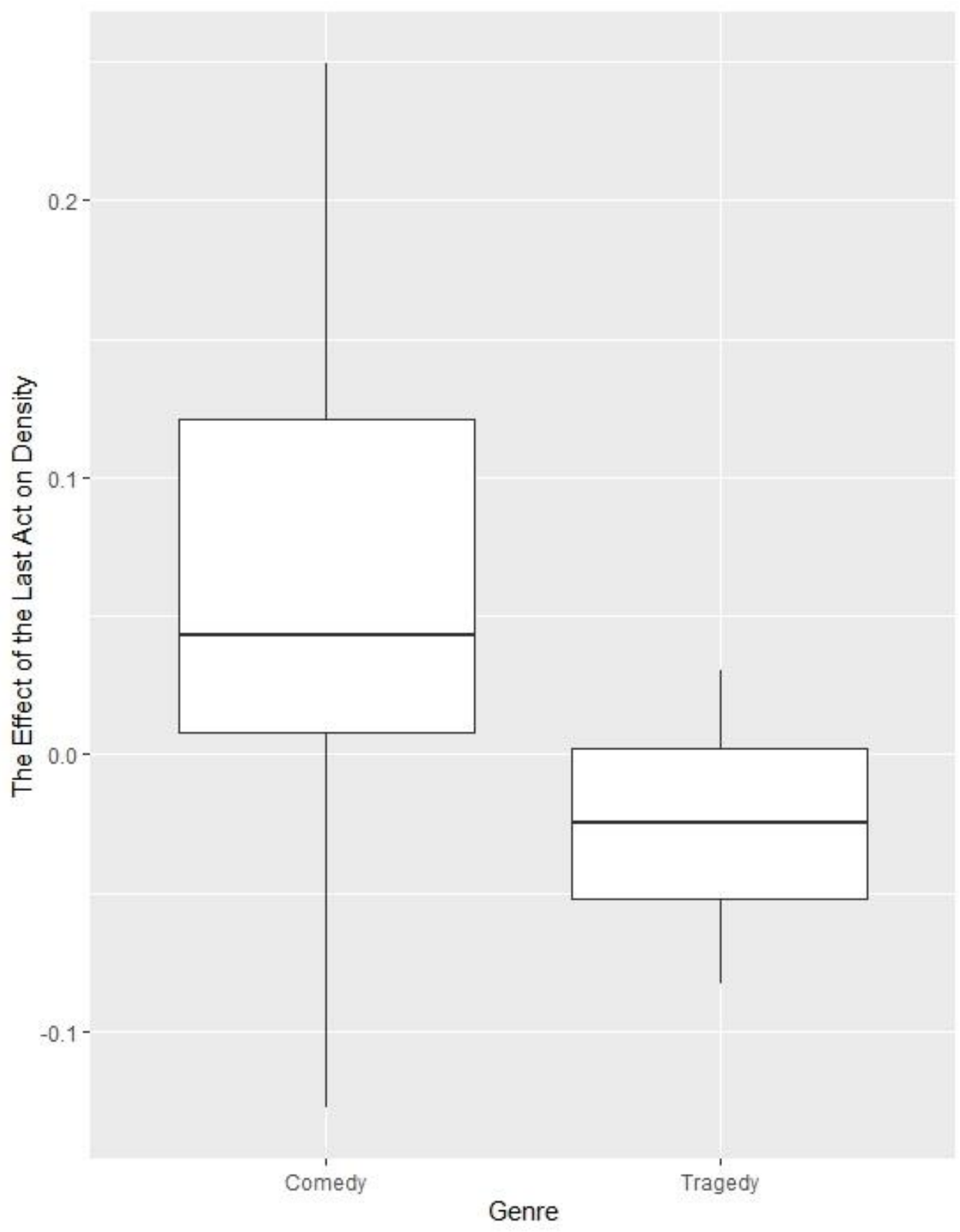

Fig6. The effect of the last act on density in the character networks of comedies and tragedies in Shakespeare. First we calculated scores for each play that show the difference between the values based on the whole drama, leaving the last act out. If the number is positive, the score for the whole play is greater than without the last act (positive effect). If the number is negative, the value of the shortened drama is greater (the ending reduces the density). We grouped the results by genre and visualized them as a boxplot.

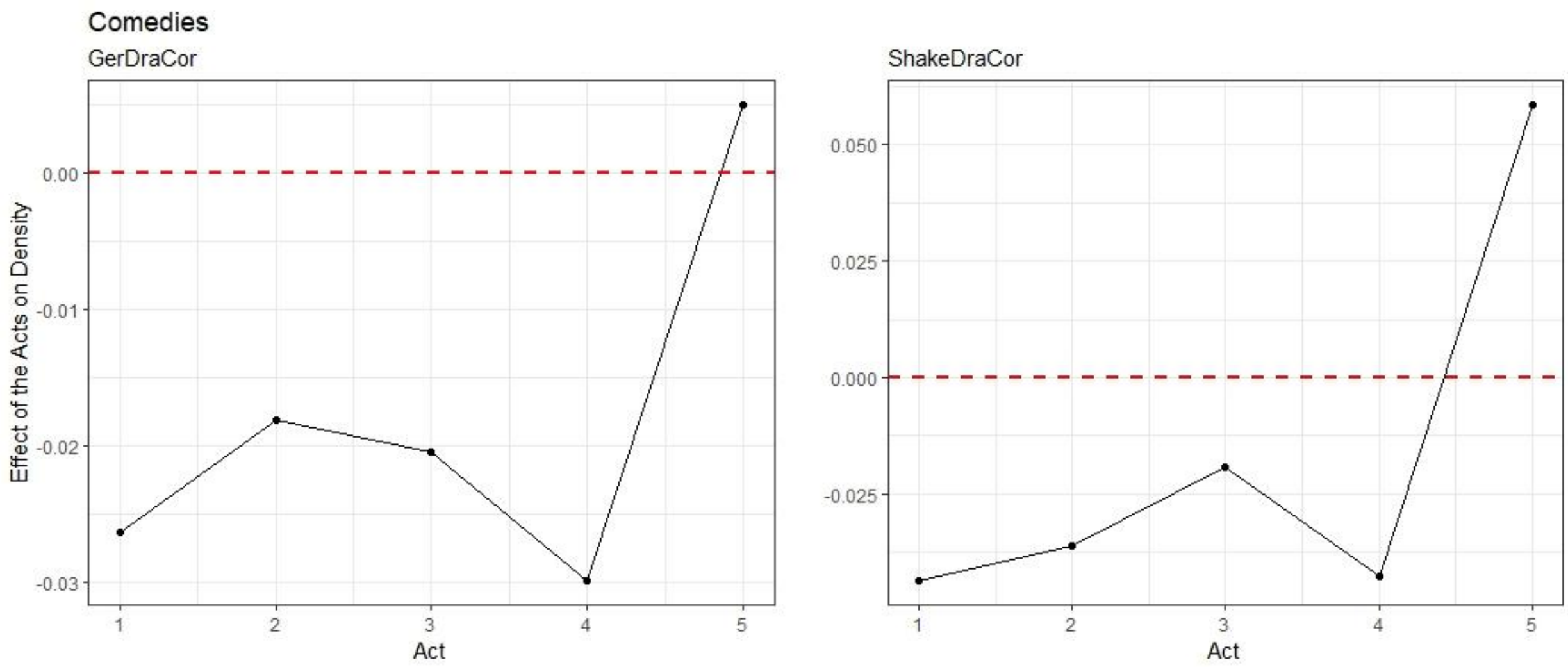

Fig7. The effect of the acts on the density of the character networks in comedies. From GerDracor we only used the five-act comedies/tragedies (128 plays)

## 5. Conclusion

It can be concluded that the measures of network theory and the distribution of speech and stage time between characters can be helpful in identifying a genre fingerprint that can be used to distinguish between comedies and tragedies. We have reached this conclusion by using a larger corpus than those of earlier relevant studies, by drawing on a number of measures and by testing their combined use, always keeping in mind that our results should be as independent as possible of the size of the networks.

This suggests that prototypical comedies are characterized by denser character networks and fewer prominent characters who control the exchange of information, talk notably more than their peers, or have significantly more connections than others. In such a network multiple pieces of information circulate simultaneously, not only leading to misunderstandings and frequent situational comedy but also preventing the will of a single character and a single worldview from prevailing in the play. The structure thus avoided is the logic of tragedies, in which networks are broken down into subgroups, one or two key figures act as the link between them, and the majority of the characters play only a peripheral role in the development of the plot (both in terms of number of links and amount of speech). While these arrangements allow for more effective communication by allowing the spread of a single or small number of truths and values (thus limiting misunderstandings and duplicities but increasing vulnerability and the possibility of deliberate deception), they are also more fragile, since the failure of this "one truth" can lead to the permanent disintegration of the whole network.

However, these genre features are characteristic of ideal types – not all comedies and tragedies follow these rules so neatly. This can be witnessed in Fig4 and Fig5, which show overlapping categories rather than a clear divide. Although the aim of all research in relevant fields of digital humanities is to develop clustering procedures that are as efficient as possible, the discipline should not abandon the investigation of categories that are based on family resemblance and prototype theory. Indeed, beside the classic approach of the philosophy of science, a principle of categorization suggested by cognitive psychology should also be considered (Rosch 1975; Rosch 1977). According to the former, "categories are defined by necessary and sufficient conditions; properties are binary (either present in the instance or not); belonging to a category is a matter of yes-no decision; categories have precise boundaries, all members of the categories are equal, the subordinate concept has all properties of the superordinate category". By contrast, cognitive psychology argues that in some cases "specimens are categorized on the basis of recognized characteristics; the main principle of classification is family resemblance (…); categorization is a matter of degree, with some specimens being central, 'good', and others less good, corresponding to fewer characteristics; the outlines of categories are not defined, they may overlap." (Tolcsvai Nagy 2005) An examination of 253 plays from the GerDracor corpus clearly suggests that prototype theory is better suited to describe the difference between comedies and non-comedies. In other words, the categories of genre classification are not absolute but part of a spectrum: even plays that are clearly considered tragedies contain features associated with comedies – such as *King Lear* and *Othello* in the Shakespeare corpus. According to the x axis in the PCA plot of Fig5, we can mention *Comedy of Errors* and *Henry V.* as examples of central members of the comedy and tragedy prototypes, respectively. The former builds up a denser network of multiple protagonists and moves more characters into the scene at once; the latter has a larger network diameter and is characterized by more supporting characters and/or characters with high betweenness centrality. This can be seen in Fig8 (a) and (b): while in *Henry V* the sub-groups of characters are linked in a chain through some central, mediating character, in *The Comedy of Errors* there is a single tightly linked group at the centre of the network, loosely connected to the more peripheral characters. Of course, the procedure presented here could be improved and replaced by a more effective model, yet the point of studies in this field – unlike, for instance authorship attribution studies – may not be the making of strict distinctions but the detection of prototypical features and the identification of instances that conform to or deviate from the scheme.

Finally, the interdependence between the plays' structure and the plot should be noted. Measurements made without the final acts have shown that the evolution of character networks in the Shakespeare corpus is influenced by the outcome of the plot (densely populated scenes of marriages vs the gradual accumulation of character deaths). In the German corpus such a strong correlation could not be observed between the parts of the plays and the change in the different features; but the relationships of the characters (whether we mean by this term the networks or the distribution of stage time and speech) in the plays cannot be considered independent of the unfolding of the story. We rather prefer to think of this as a reciprocal connection. It cannot be argued that it is the structural features of comedies and tragedies that create comic and tragic plots; neither can it be argued, that the characters' system of relations is determined exclusively by the plot development. Rather, there is an interaction between the two sides; certain stories can unfold within a certain framework, but in doing so they also influence the structure of that framework. There have also been examples of a plot of one genre combining with the structure of another genre, in which case it is the examination of the resulting tension that can lead to important insights.

In the future it is worth adding another aspect of temporality to these results; are genre differences considered to be constant over time, or do these differences evolve historically? With such a study it would then become possible to examine how dramatic forms change and are preserved within a framework of cultural evolution.[10]

a.)

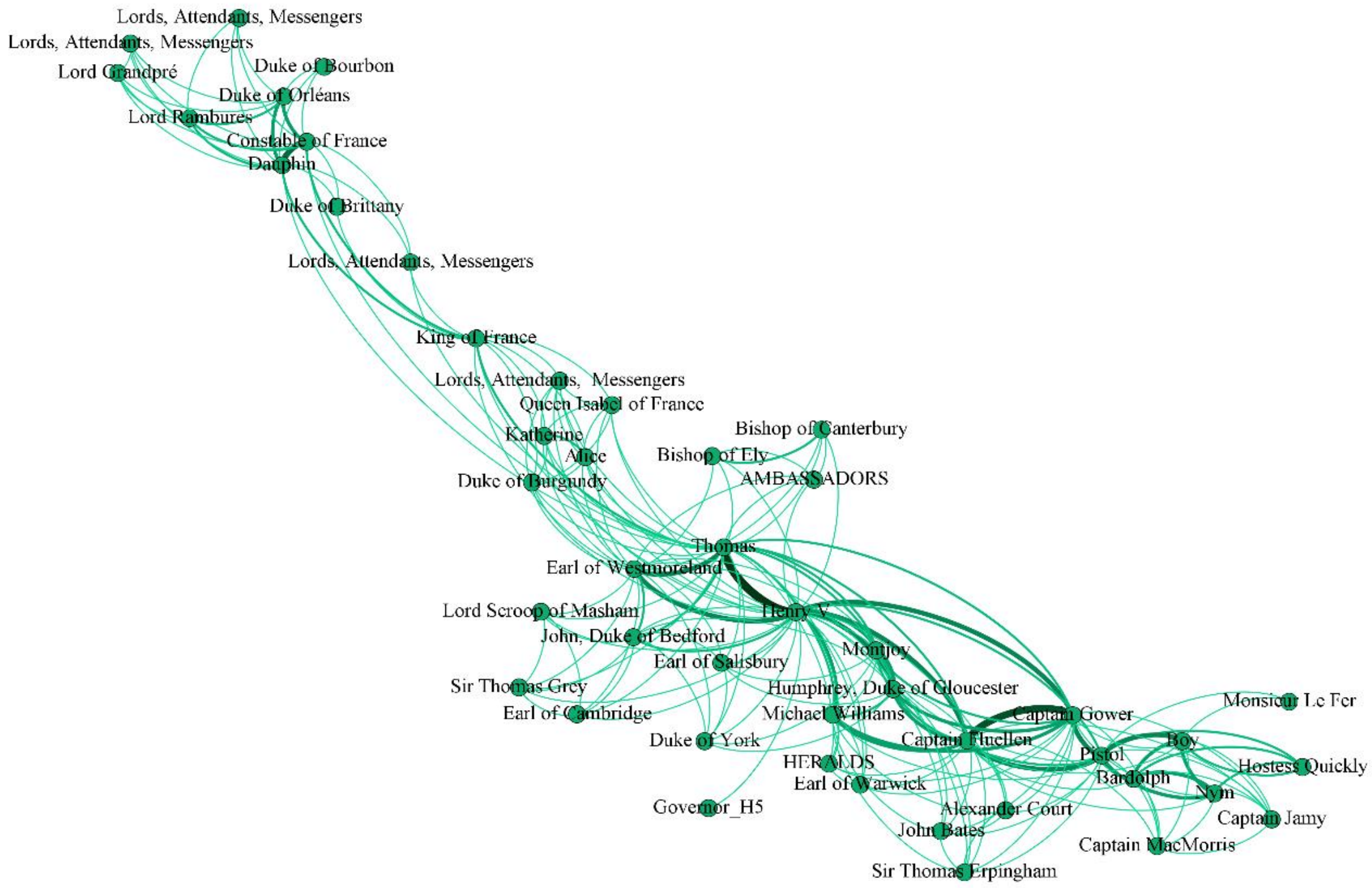

10 This discipline studies the change, persistence and diffusion of information acquired through social learning - that is, it considers cultural processes to be captured along the dynamics of innovation and preservation. To model this, it draws on the insights and terminology of evolutionary theory. (see eg. Mesoudi 2017)

b.)

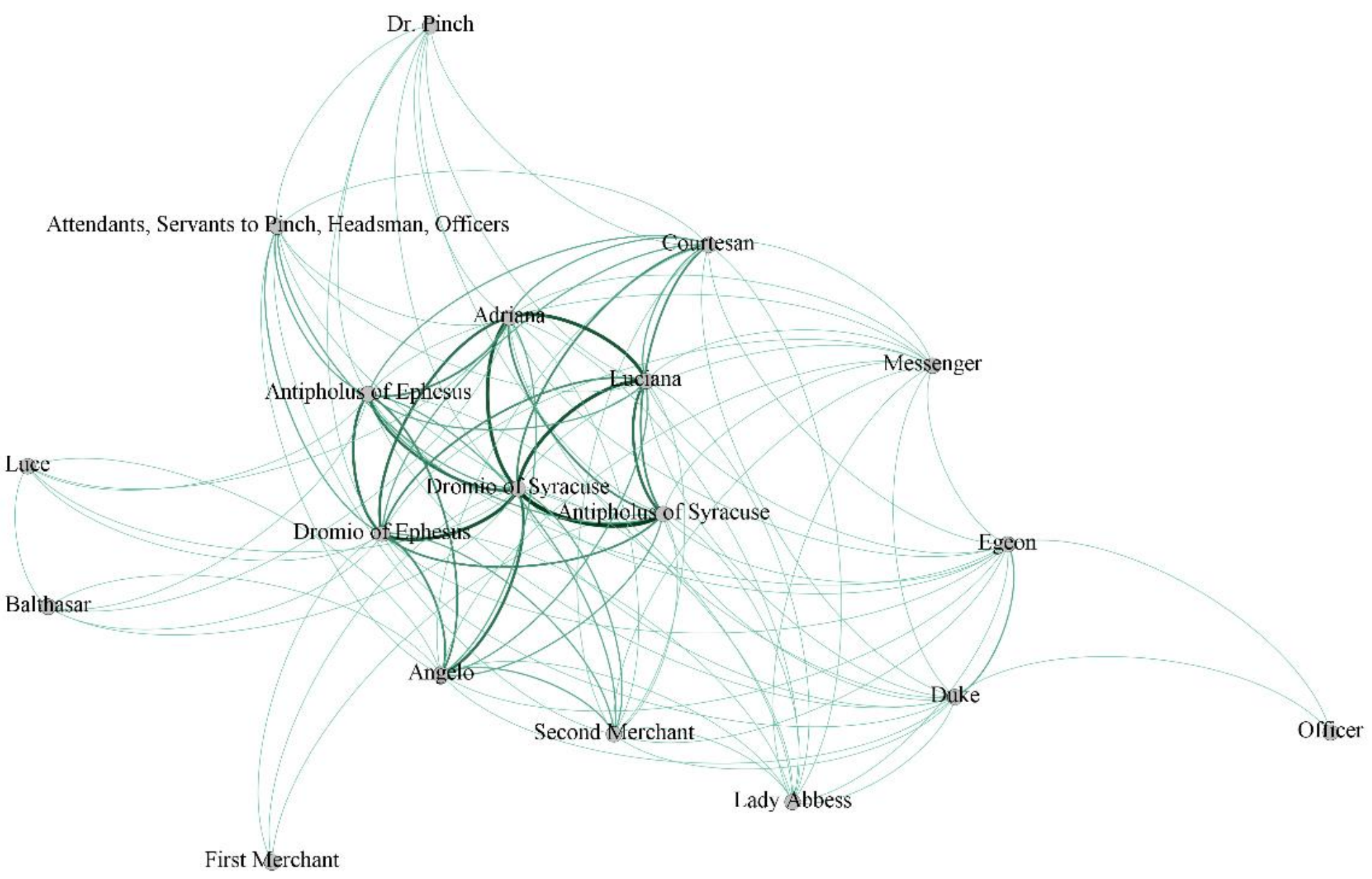

Fig7. Weighted character networks based on co-occurrence in a scene. Network data was downloaded from the Dracor website and processed in the Gephi software using the ForceAtlas2 algorithm – a) *Henry V.*, b) *Comedy of Errors*.

## Data availability

The data extracted from the DraCor database was analyzed in the R programming environment; the data for testing the effect of the acts on the network structure (4.1.) was produced in python – the codes and the results were made publicly available in the following GitHub repository: https://github.com/SzemesBotond/drama_cluster_genre

## Acknowledgement

Szemes Botond was supported by the ÚNKP-22-4 New National Excellence Program of the Ministry for Culture and Innovation (Hungary) from the source of the National Research, Development and Innovation Fund.

The authors are grateful for the help of Mihály Nagy (ELTE University) in the computations of the network metrics without given acts; and Jessie Labov (Research Centre for the Humanities, Institute for Literary Studies Budapest) in proof reading the text.